\documentclass{article}
\usepackage{amsmath,amssymb}
\usepackage{graphicx}
\usepackage{booktabs}
\usepackage{siunitx}
\usepackage{hyperref}
\usepackage[nameinlink,noabbrev]{cleveref}
\usepackage{caption}
\usepackage{subcaption}
\newcommand{\includefig}[2][]{%
  \IfFileExists{#2}{\includegraphics[#1]{#2}}{%
    \fbox{%
      \parbox[c][0.22\textheight][c]{0.95\linewidth}{\centering\small Missing figure file:\\\texttt{\detokenize{#2}}}%
    }%
  }%
}

\usepackage{arxiv}
\usepackage[T1]{fontenc}
\usepackage[sc]{mathpazo}
\usepackage{microtype}
\usepackage[numbers,sort&compress]{natbib}
\title{Learning Response-Statistic Shifts and Parametric Roll Episodes from Wave--Vessel Time Series via LSTM Functional Models}

\author{Jos\'e del \'Aguila Ferrandis\thanks{Corresponding author. Email: \texttt{jaguila@vt.edu}}\\
Kevin T. Crofton Department of Aerospace and Ocean Engineering, Virginia Tech, USA}

\begin{document}
\maketitle

\begin{abstract}
Parametric roll is a rare but high-consequence instability that can trigger abrupt regime changes in ship response, including pronounced shifts in roll statistics and tail risk.
This paper develops a data-driven surrogate that learns the nonlinear, causal functional mapping from incident wave--motion time series to vessel motions, and demonstrates that the surrogate reproduces both (i) \emph{parametric roll episodes} and (ii) the associated \emph{statistical shifts} in the response.
Crucially, the learning framework is \emph{data-source agnostic}: the paired wave--motion time series can be obtained from controlled experiments (e.g., towing-tank or basin tests with wave probes and motion tracking) when a hull exists, or from high-fidelity simulations during design when experiments are not yet available.
To provide a controlled severe-sea demonstration, we generate training data with a URANS numerical wave tank, using long-crested irregular seas synthesized from a modified Pierson--Moskowitz spectrum.
The demonstration dataset comprises 49 random-phase realizations for each of three sea states, simulated at a fixed forward speed selected to yield encounter conditions under which parametric-roll episodes can occur.
A stacked LSTM surrogate is trained on wave-elevation time series and evaluated on held-out realizations using time-domain accuracy and distributional fidelity metrics.
In the most severe case, the model tracks the onset and growth of large-amplitude roll consistent with parametric excitation, and captures the corresponding changes in roll probability density functions (PDFs).
We further compare loss-function choices (MSE, relative-entropy-based objectives, and amplitude-weighted variants) and show how they trade average error for improved tail fidelity relevant to operability and risk assessment.
\end{abstract}


\section{Introduction}
Extreme-sea operability and safety are often governed by \emph{rare, nonlinear response events} rather than by average motion levels.
A prominent example is \emph{parametric rolling}, in which periodic variations in restoring characteristics and encounter conditions can excite subharmonic roll growth, producing intermittent large-amplitude roll, abrupt changes in response regime, and pronounced shifts in response statistics \citep{paulling1959unstable,france2003investigation,neves2006unstable,silva2005parametrically}.
These events challenge linear or weakly nonlinear seakeeping assumptions precisely in the operating regimes where accurate prediction matters most.
They are also expensive to quantify with high-fidelity CFD, since reliable estimates of exceedance probabilities, PDFs, or other tail-oriented metrics require many long irregular-sea realizations.

Machine-learning surrogates have therefore become increasingly attractive for accelerating the exploration of complex nonlinear physics, including multi-fidelity modeling with Gaussian processes and neural networks \citep{perdikaris2015multi,perdikaris2017nonlinear,bonfiglio2018probabilistic,meng2020composite}, and broader deep-learning advances in fluid dynamics and nonlinear systems \citep{ling2016reynolds,kutz2017deep}.
Within marine hydrodynamics specifically, recent studies have already shown that recurrent and related neural architectures can learn wave--motion relationships from high-fidelity simulations or measurements.
Examples include functional LSTM surrogates for vessel motions in extreme seas \citep{RoyProccA_JAF}, recurrent models for nonlinear marine dynamics driven by wave probes or wave-elevation histories \citep{xu2021nonlinearmarine}, data-driven system identification of 6-DoF ship motion in irregular waves \citep{silva2022sixdof}, very-short-term roll prediction enhanced by wave-elevation input \citep{tian2023shortterm}, multi-step roll prediction in high sea states \citep{zhang2023multistep}, and recent spatiotemporal wave-field-based neural networks for nonlinear ship motions including parametric roll \citep{lee2025spatiotemporal}.

In parallel, a complementary body of naval-architecture research has focused directly on \emph{extreme-response characterization}.
This includes Critical Wave Groups (CWG)-based probabilistic frameworks and temporal exceedance metrics for rare ship-response events \citep{silva2023cwg,gong2022temporal}, as well as analytical and stochastic approaches for non-Gaussian roll statistics and parametric-roll amplitude distributions \citep{maki2019nongaussian,maruyama2024pdf}.
These studies are important because, from an operability and risk perspective, the quantity of interest is often not only a time trace or a mean-square error, but the \emph{change in the response distribution} and the corresponding tail behavior.

Against this background, the question addressed here is deliberately focused: can one learn a nonlinear map from \emph{incident wave histories} to \emph{motion histories} that reproduces both parametric-roll episodes and the associated shift in response statistics?
We adopt a \emph{functional} viewpoint, seeking a nonlinear map from input time histories to response time histories rather than a static input--output relation.
Universal approximation results for continuous functionals \citep{Functionals} motivate this perspective: the object to be learned is a map between histories.
We represent this functional with stacked Long Short-Term Memory (LSTM) networks \citep{LSTM}, whose temporal memory is naturally suited to resonance build-up, delayed nonlinear effects, and intermittent growth and decay.

The most closely related learning-based parametric-roll studies motivate, but do not fully resolve, this question.
Zhou et al.~\citep{zhou2023parametricroll} propose a grey-box strategy that combines a low-order parametric-roll equation with learning components and a recognition model for wave histories that trigger large-amplitude events.
Silva and Maki \citep{silva2023cwg} embed LSTM surrogates within the CWG framework to accelerate exceedance-probability estimation for extreme motions.
More recently, Lee and Kim \citep{lee2025spatiotemporal} demonstrated that spatiotemporal wave-field-based neural networks can predict nonlinear ship motions, including parametric roll, and assess the predictions from deterministic, probabilistic, and statistical viewpoints.
However, these studies do not center the problem on \emph{jointly} reproducing regime transitions in irregular seas and preserving the induced response-distribution shift and tail behavior within an end-to-end wave-history-to-motion surrogate.

The formulation considered here is intentionally \emph{data-source agnostic}: the required paired wave--motion time series may be obtained from controlled experiments when a hull exists, or from high-fidelity simulations during design.
We view this flexibility as a practical advantage rather than as the primary scientific novelty.
The novelty of the present paper lies instead in the combination of five elements:
(i) end-to-end functional learning from incident wave histories to response histories;
(ii) a focus on parametric-roll regime transitions in irregular severe seas;
(iii) explicit evaluation of \emph{distributional fidelity}---including PDF shifts and tail behavior---rather than relying only on trajectory-wise error; and
(iv) a targeted study of loss-function design to improve tail preservation relevant to operability and risk assessment. (v) the release of a curated severe-sea URANS wave--motion dataset designed to exhibit regime-dependent nonlinear roll behavior, thereby providing a reference for future learning-based studies of extreme ship response.

\paragraph{Contributions}
The paper is organized around the following contributions:
\begin{enumerate}
  \item \textbf{End-to-end nonlinear functional surrogate:} we formulate wave-to-motion prediction as nonlinear functional approximation using stacked LSTM networks. The formulation applies to paired wave--motion time series obtained either experimentally or numerically, enabling the learning of strongly nonlinear responses as functionals of the incident-wave history.
  \item \textbf{Controlled severe-sea demonstration:} to provide a repeatable and physically interpretable testbed, we use an ensemble URANS numerical wave-tank campaign in long-crested irregular seas across three sea states (49 random-phase realizations per sea state), under encounter conditions chosen so that a subset of realizations exhibits strong nonlinear roll behavior and regime-dependent statistics.
  \item \textbf{Regime-transition fidelity:} we show that the learned surrogate can reproduce the onset and growth of parametric-roll-like episodes on held-out irregular-sea realizations.
  \item \textbf{Distribution-shift and tail fidelity:} we show that the surrogate captures changes in response statistics, especially in the roll PDF and tails, and we analyze how loss-function choices (MSE, relative-entropy-based, and amplitude-weighted objectives) trade average trajectory error against tail fidelity relevant to operability and risk.
  \item \textbf{Open severe-sea benchmark dataset:} we release the curated URANS wave--motion dataset used in this study, to provide a reference testbed for future learning-based studies of nonlinear ship response and parametric roll.

\end{enumerate}

In this sense, the paper complements prior ML work on ship-motion prediction and prior extreme-response methods for roll statistics.
Rather than proposing another generic ship-motion predictor, it asks whether an end-to-end learned functional can preserve the specific regime changes and statistical shifts that matter for extreme-sea decision-making.

\section{Data sources and demonstration dataset}
The simulated conditions correspond to the DTMB 5415 advancing at a Froude number of $Fr=0.4$ in oblique irregular seas, with waves incident at $30^\circ$ relative to the vessel longitudinal axis. This operating point is chosen because it yields encounter conditions under which strong nonlinear roll behavior, including parametric-roll episodes, can arise while remaining representative of a practically relevant seakeeping scenario \cite{delaguila2026thesis}.

\subsection{Sea-state definition and irregular-wave synthesis}
Long-crested irregular seas are represented by a superposition of harmonic components with random phases,
\begin{equation}
  \zeta(t)=\sum_{i=1}^{n} a_i \cos\!\left(\omega_i t + \varepsilon_i\right),
\end{equation}
where amplitudes $a_i$ are selected to match a target spectral density.
We employ a modified Pierson--Moskowitz spectrum \citep{PMS}.
The dataset comprises 49 random-phase realizations for each of three sea states (Table~\ref{tab:seastates}).

Each sea state comprises 49 independent random-phase realizations, with 40 used for training and 9 held out for testing, and a pooled simulated duration of slightly more than one hour per ensemble.
In seakeeping practice, statistical adequacy is often discussed in terms of the number of wave encounters rather than wall-clock time; rule-of-thumb guidance frequently targets $O(10^2)$ encounters for stable bulk statistics \citep{ittc29_seakeeping_2021}.
Using the peak periods in Table~\ref{tab:seastates} as a dominant scale gives $N \approx T_{\mathrm{tot}}/T_p \approx 371$ (SS--1), $290$ (SS--2), and $269$ (SS--3) dominant cycles over the pooled ensembles, which supports convergence of core moments and the central portion of the roll PDF.
However, parametric roll is episodic and exhibits memory through resonance build-up/decay; therefore, tail-sensitive metrics have a smaller effective sample size than the raw wave count.
To remain conservative, we interpret tail-related conclusions primarily in terms of robust distributional \emph{shifts} (Gaussian-like in SS--1/SS--2 versus heavy-tailed in SS--3) rather than claiming tight convergence of very low exceedance probabilities, and we report tail-fidelity comparisons across loss functions under this practical sampling constraint.

\begin{table}[t]
\centering
\caption{Sea states in the URANS campaign (49 random-phase seeds per sea state). Across the full campaign (three sea states, 49 random-phase realizations each), the total simulated duration of each ensemble exceeds \SI{3600}{s}.}

\label{tab:seastates}
\begin{tabular}{@{}lcc@{}}
\toprule
Case & Significant height $H_s$ [m] & Peak period $T_p$ [s] \\
\midrule
SS--1 & 3.53 & 9.7 \\
SS--2 & 5.09 & 12.4 \\
SS--3 & 10.66 & 13.4 \\
\bottomrule
\end{tabular}
\end{table}
\subsection{URANS solver and recorded signals}
Disregarding viscous effects can substantially underestimate energy dissipation for surface-piercing bodies, and this limitation is particularly acute for rolling responses in steep, nonlinear seas and near resonance, where motions are large relative to the characteristic dimensions.
Because the objective of this work is to learn strongly nonlinear phenomena (including parametric-roll regime transitions and the associated distribution shifts), we generate a demonstration dataset using a viscous, free-surface solver that resolves both nonlinear wave kinematics and viscous dissipation mechanisms.
Specifically, we employ an incompressible Unsteady Reynolds-Averaged Navier--Stokes (URANS) framework with empirical turbulence closure \citep{ferziger2012computationa,itemreference3,ruth2015simulation,shen2013rans,Stern1,Stern2,menter1994two}. The URANS dataset is obtained from a viscous VOF solver. The averaged continuity and momentum equations for incompressible flow can be written as
\begin{align}
\frac{\partial \overline{u}_{i}}{\partial x_{i}} &= 0, \label{eq:urans_cont}\\
\frac{\partial \overline{u}_{i}}{\partial t} +
\frac{\partial}{\partial x_{j}}\left(\overline{u}_{i}\overline{u}_{j}+\overline{u_{i}'u_{j}'}\right)
&=
-\frac{1}{\rho}\frac{\partial \overline{p}}{\partial x_{i}}
+\frac{1}{\rho}\frac{\partial \overline{\tau}_{ij}}{\partial x_{j}}, \label{eq:urans_mom}
\end{align}
with viscous stress tensor
\begin{equation}
\overline{\tau}_{ij}=\mu\left(\frac{\partial \overline{u}_{i}}{\partial x_{j}}+\frac{\partial \overline{u}_{j}}{\partial x_{i}}\right), \label{eq:urans_tau}
\end{equation}
where $\overline{p}$ is the averaged pressure, $\overline{u}_i$ are the components of averaged velocity, $\overline{u'_i u'_j}$ are Reynolds stresses, and $\rho$ and $\mu$ are fluid density and dynamic viscosity.
Turbulence closure is provided by an empirical URANS model \citep{menter1994two}.
The free surface is modeled using a VOF method \citep{VOF}, which advects a phase fraction and applies the governing equations across phases using mixture properties (see \citep{ferziger2012computationa}).
Rigid-body motions are coupled through DFBI, allowing heave, pitch, and roll as required by the case.


The air--water interface is captured with a Volume-of-Fluid (VOF) method \citep{VOF}, in which a phase volume fraction is advected and the same governing equations are applied across phases with appropriately defined mixture properties (see, e.g., \citep{ferziger2012computationa}).
Vessel motions are obtained through a Dynamic Fluid--Body Interaction (DFBI) coupling, treating the hull as a rigid body allowed to move in the relevant degrees of freedom.
In this paper we consider configurations with three DOFs (heave, pitch, and roll), depending on the case, enabling the emergence of large-amplitude roll dynamics under severe-sea excitation. For each realization we record sea-surface elevation time series (at one or more probes) together with the vessel motions (heave, pitch, roll), which form the paired inputs/outputs for the LSTM functional surrogate.


\section{Functional surrogate model}
\subsection{Problem formulation and spatiotemporal input stencil}
Much like in numerical methods, the \emph{representation} used to feed data into a learning algorithm can emphasize or obscure physically relevant structure.
In PDE discretization, this is controlled by the numerical stencil (e.g., centered vs upwind differences), which trades accuracy, stability, and cost \citep{leveque2002finite,godunov1959difference}.
Larger stencils can represent richer dynamics (e.g., higher-frequency content), but they can also amplify numerical artifacts and accumulated errors, effectively injecting ``noise'' \citep{chapra2010numerical,trefethen2022numerical}.
A closely related tradeoff appears in machine learning: inputs that are too impoverished under-represent the physics, while overly large or weakly relevant inputs increase variance and can degrade generalization by forcing the model to learn an implicit filtering task \citep{burden2011numerical,kotsiantis2006machine,domingos2012few}.

\begin{figure}[t]
\centering

\begin{subfigure}[t]{0.48\columnwidth}
  \vspace{0pt}
  \centering
  \frame{\includegraphics[width=\linewidth]{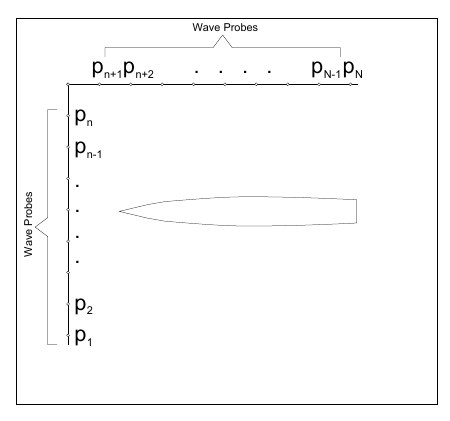}}
  \caption{Schematic of wave probes in the computational domain.}
  \label{fig:SchematicWaveProves}
\end{subfigure}
\hfill
\begin{subfigure}[t]{0.48\columnwidth}
  \vspace{0pt}
  \centering
  \includegraphics[width=\linewidth]{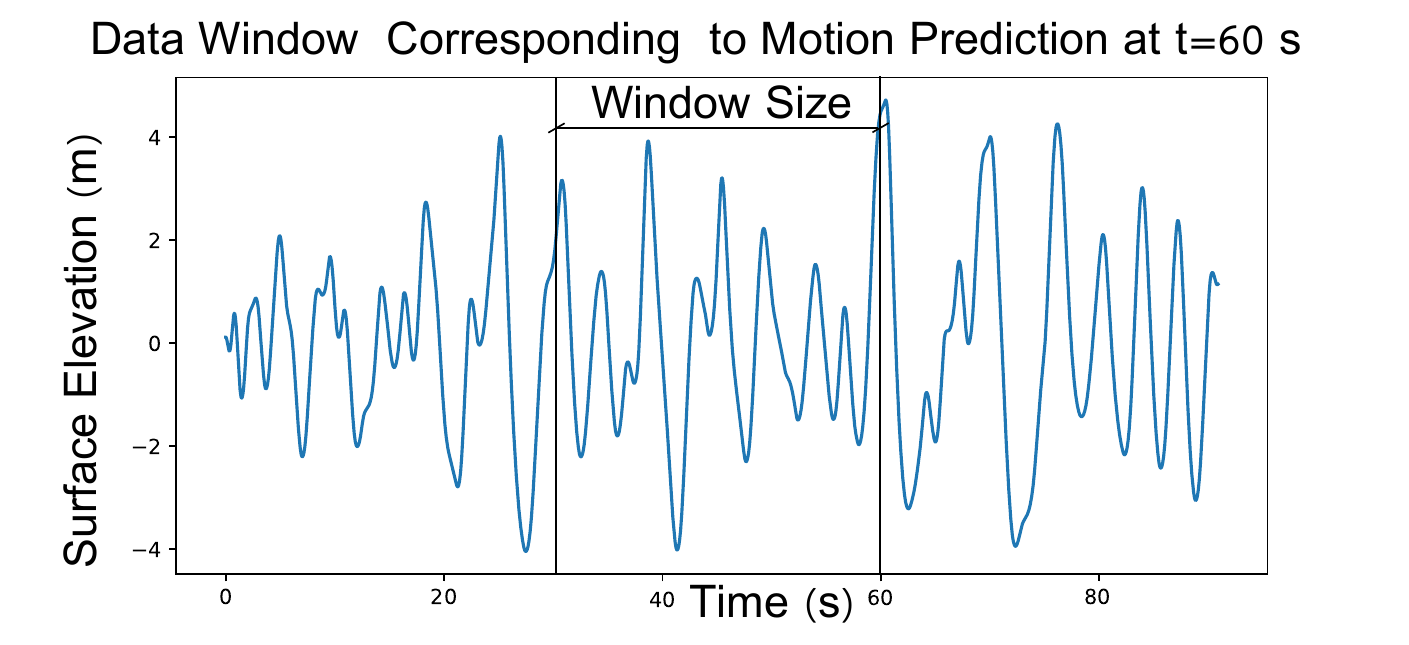}

  \vspace{0.6em}

  \includegraphics[width=\linewidth]{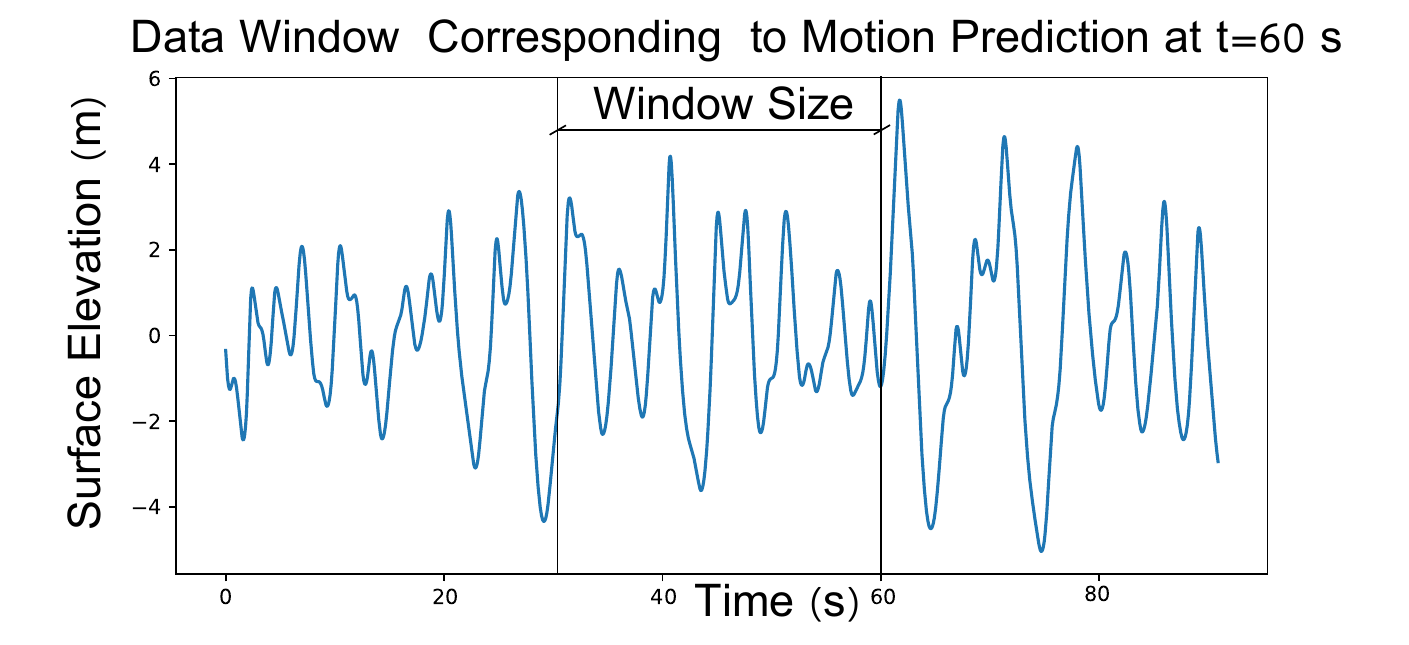}
  \caption{Wave profile in the longitudinal (top) and transversal (bottom) directions for test case one.}
  \label{fig:Window}
\end{subfigure}

\caption{Inputs for the spatiotemporal stencil: (left) probe layout; (right) representative wave profiles in longitudinal and transversal directions.}
\label{fig:InputsStencil}
\end{figure}

We therefore design a \emph{spatiotemporal stencil} for wave--motion learning (see \cref{fig:SchematicWaveProves,fig:Window}).
Let the vessel motions be
\begin{equation}
  y(t) \equiv \big(\eta(t),\,\theta(t),\,\phi(t)\big),
\end{equation}
where $\eta$ is heave, $\theta$ is pitch, and $\phi$ is roll.
Let $p_n(t)$ denote the measured (or simulated) free-surface elevation at wave probe $n\in\{1,\dots,N\}$.
To predict motions at time $t$, we provide the network with a \emph{causal history window} of each probe,
\begin{equation}
  P_n^{K}(t) \equiv \big[p_n(t),\,p_n(t-\Delta t),\,\dots,\,p_n(t-K\Delta t)\big]^\top \in \mathbb{R}^{K+1},
  \label{eq:probe_history}
\end{equation}
and stack across probes to form the input stencil
\begin{equation}
  X(t) \equiv \big[P_1^{K}(t),\,P_2^{K}(t),\,\dots,\,P_N^{K}(t)\big] \in \mathbb{R}^{(K+1)\times N}.
  \label{eq:stencil_input}
\end{equation}
We then learn the wave-to-motion map as
\begin{equation}
  y(t) = \mathcal{G}\big(X(t)\big) \approx f_{\mathrm{NN}}\!\big(X(t)\big),
  \label{eq:nn_map}
\end{equation}
where $f_{\mathrm{NN}}$ is implemented with a recurrent architecture (stacked LSTMs in this work) to exploit temporal memory and to represent delayed nonlinear effects.

The stencil parameters $N$ (number/placement of probes), $\Delta t$ (sampling interval), and $K$ (history length) play roles analogous to stencil width in numerical methods.
Choosing $K$ too small can omit physically relevant memory (e.g., resonance build-up and intermittent growth/decay associated with parametric roll), while choosing $K$ too large can introduce weakly relevant or redundant information that burdens learning and can reduce robustness.
Accordingly, $K$ and related architectural choices are treated as hyperparameters and selected via validation/architecture-convergence studies.
This formulation is also data-source agnostic: $p_n(t)$ can come from experiments (wave probes and motion tracking) or from high-fidelity simulations, and the learned functional can be evaluated consistently under either setting.

\subsection{Stacked LSTM surrogate}
We approximate $\mathcal{G}$ with a stacked LSTM followed by a linear readout.
Unless otherwise stated, we adopt the same LSTM architecture and training protocol used in our prior functional-learning study \citep{RoyProccA_JAF} to isolate the effect of the evaluation setting (regime transitions and distribution shifts) rather than architectural novelty.
Inputs are spatiotemporal windows of wave elevation, e.g.,
$\{x(t-k\Delta t)\}_{k=0}^{K}$ (or the multi-probe stencil in \cref{eq:stencil_input}), and targets are the corresponding vessel motions $y(t)$.

Importantly, the input wave histories are \emph{not} augmented with explicit sea-state labels or parameters (e.g., $H_s$, $T_p$).
That is, we do not condition the model on sea-state descriptors; instead, the network must infer the prevailing forcing regime directly from the raw elevation history.
This design choice is deliberate: conditioning on $(H_s,T_p)$ would simplify the learning problem, whereas our objective is to assess whether an end-to-end functional surrogate can \emph{track shifts in response statistics} (including roll PDF/tail changes) as sea-state characteristics vary, using only time-local wave information.
Architectural choices (number of layers/hidden units and window length $K$) are evaluated via held-out realizations. In particular, when training on pooled realizations from multiple sea states, the same model is exposed to distinct response regimes without any explicit regime identifier.

\subsection{Loss functions emphasizing tail fidelity}
For severe-sea operability, the cost of \emph{underpredicting} large responses can exceed the cost of small average errors, because risk metrics depend on the tail of the response distribution (e.g., exceedance probabilities).
Accordingly, in addition to standard mean-square objectives, we consider losses that (i) emphasize extremes and (ii) better preserve distributional properties.
In our experiments, plain MSE typically yields the best average time-domain accuracy, but can underrepresent rare large-amplitude events; the alternative losses below trade some average accuracy for improved tail fidelity.

\subsubsection{Mean square error (MSE)}
We use the standard mean square error
\begin{equation}
\mathrm{MSE}=\frac{1}{n}\sum_{i=1}^{n}\left(Y_i-\hat{Y}_i\right)^2,
\label{eq:mse}
\end{equation}
where $Y_i$ and $\hat Y_i$ denote reference and predicted responses.
MSE is a strong baseline for trajectory fidelity and, through its quadratic penalty, it already assigns increased weight to large pointwise deviations.
However, it remains a \emph{sample-averaged, pointwise, symmetric} objective: when extreme responses occupy a small fraction of the time series or of the realization ensemble, the optimization can still be dominated by the high-probability ``core'' of the distribution.
As a result, low MSE does not necessarily imply accurate preservation of tail-sensitive distributional quantities (e.g., exceedance probabilities, tail mass, or regime-dependent shifts in the roll PDF).
This motivates losses that \emph{further concentrate} the learning signal on extreme-amplitude events and/or more directly promote distributional fidelity in the tails.

To operationalize this goal we consider two complementary modifications:
(i) a \emph{relative-entropy-inspired} objective that reweights samples via an exponential-tilting construction, so that large-amplitude outcomes contribute disproportionately to the loss.
Because the tilting induces an auxiliary distribution that concentrates on extremes, minimizing a KL divergence between the tilted ``truth'' and tilted ``prediction'' more directly targets tail-relevant mass than MSE, while still being trained from time-series samples.
(ii) an \emph{amplitude-weighted MSE} that retains the simplicity and stability of quadratic regression but introduces an explicit weighting $w(Y)$ that grows with $|Y|$; this provides a transparent knob for prioritizing peak-amplitude/episode fidelity and can improve tail behavior when extremes are underrepresented in the unweighted sample average.

\begin{figure}[h!]
\centering

  \includefig[width=\textwidth]{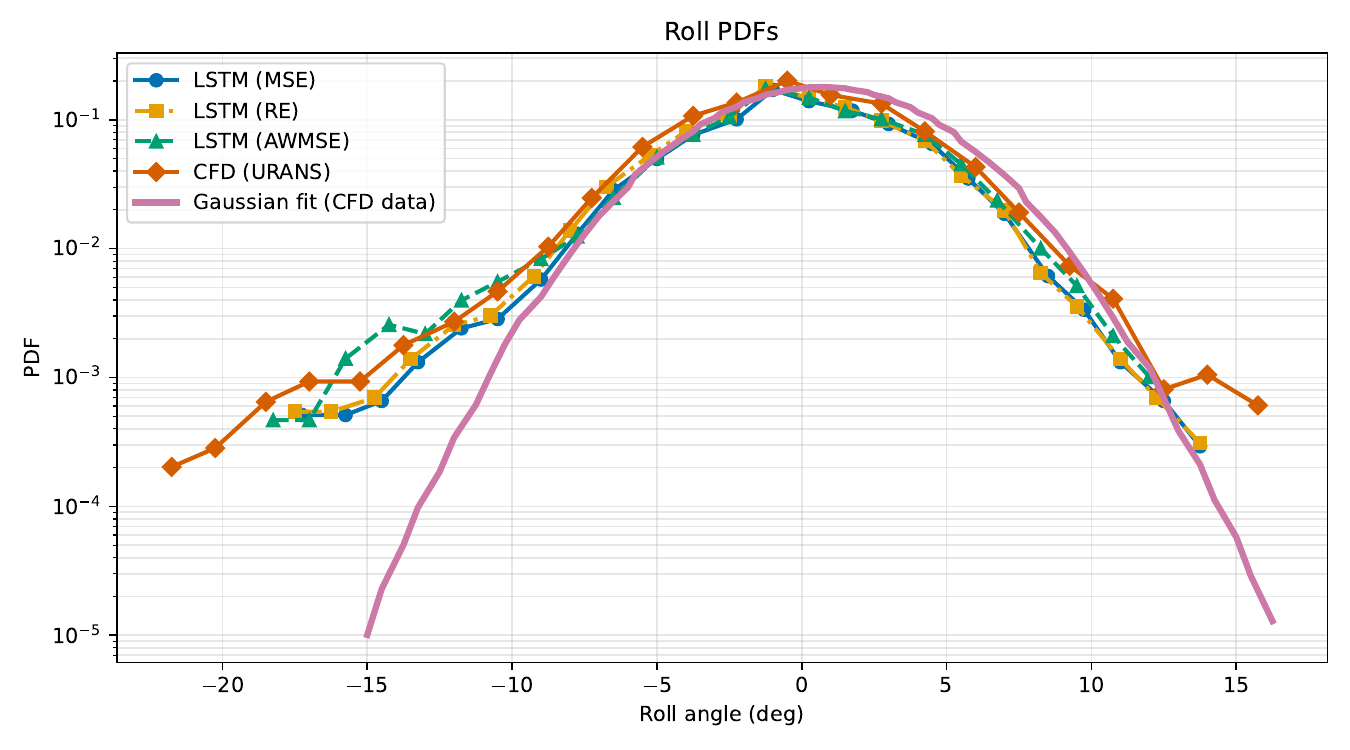}

\caption{Roll response-statistics fidelity in the most severe sea state (SS--3): comparison of observed URANS roll PDF to learned surrogate-induced PDFs under different training objectives. To show the departure from gaussian statistics we include a fit with a gaussian distribution of the URANS roll PDF.}
\label{fig:comp_pdf_134}
\end{figure}

\subsubsection{Relative-entropy-inspired tail emphasis for scalar outputs}

Relative entropy (KL divergence) has been used to emphasize extreme-event prediction in dynamical systems \citep{RUDY2023133570,qi2020using}.
Qi and Majda \citep{qi2020using} construct a KL-based loss by applying a softmax operator across \emph{output dimensions}, which naturally highlights large-magnitude features in high-dimensional fields.
For scalar responses (e.g., roll angle at a given time), a direct softmax across dimensions is not meaningful; we therefore adopt an analogous idea by applying an \emph{exponential tilting} (softmax-like) normalization across samples, which concentrates weight on large-amplitude outcomes.
Exponential tilting is closely related to importance-sampling constructions \citep{siegmund1976importance}.

Let $f(x)$ denote the true scalar response as a function of input $x$ (e.g., the wave-history stencil) and $\hat f(x)$ the model prediction.
Define the exponential-tilting operator
\begin{equation}
G(f)(x)=\frac{e^{f(x)}\,p_x(x)}{\mathbb{E}_x\!\left[e^{f(x)}\right]},
\label{eq:exp_tilt}
\end{equation}
which induces a density over input samples that upweights inputs associated with large responses.
We then define a relative-entropy loss as
\begin{equation}
L_{\mathrm{RE}}(\hat f)=\mathrm{KL}\!\left(G(f)\,\Vert\,G(\hat f)\right).
\label{eq:re_def}
\end{equation}
In practice, we minimize a computable surrogate objective obtained by replacing expectations with empirical averages over a batch $\mathcal{D}$ (and using standard numerical-stability tricks such as log-sum-exp for exponentials):
\begin{equation}
L_{\mathrm{RE}}(\hat f)\;\approx\;\mathbb{E}_{\mathcal{D}}\!\left[e^{\hat f(x)}-e^{f(x)}\,\hat f(x)\right],
\label{eq:re_empirical}
\end{equation}
which retains the desired effect of exponentially emphasizing large responses.
Following \citep{qi2020using}, we also include a small penalty for negative outliers by adding a ``mirror'' term based on $-f$,
\begin{equation}
L_{\mathrm{RE},\lambda}(\hat f)=L_{\mathrm{RE}}(\hat f)+\lambda\,L_{\mathrm{RE}}^{(-)}(\hat f),
\label{eq:re_lambda}
\end{equation}
where $L_{\mathrm{RE}}^{(-)}$ is defined by replacing $(f,\hat f)$ with $(-f,-\hat f)$ in \cref{eq:re_empirical}.
A small $\lambda$ biases learning toward improved fidelity for the dominant (often positive) extremes while still discouraging pathological behavior in the opposite tail.

The full derivation and bounding steps that motivate the practical form \cref{eq:re_empirical} follow the same philosophy as KL-based extreme-event training in \citep{qi2020using,RUDY2023133570} and the exponential-tilting/importance-sampling literature \citep{efron1981nonparametric,siegmund1976importance}; for brevity we omit the longer derivation here.

\subsubsection{Amplitude-weighted mean square error (AWMSE)}
As a simpler alternative to KL-style objectives, we also consider an amplitude-weighted MSE,
\begin{equation}
\mathrm{AWMSE}=\frac{1}{n}\sum_{i=1}^{n}\left(Y_i-\hat{Y}_i\right)^2\,w(Y_i),
\qquad
w(Y)=1+\beta\left(\frac{Y}{\sigma_Y}\right)^2,\qquad \beta>0
\label{eq:awmse}
\end{equation}

which increases the penalty on errors at large amplitudes.
In implementation, $w(Y)$ must be kept nonnegative (e.g., by choosing $(\epsilon_1,\epsilon_2)$ such that $w(Y)\ge 0$ over the training range, or by clipping), to avoid unstable optimization.
AWMSE provides a direct knob for penalizing underperformance on peaks, at the cost of potentially increased sensitivity to outliers.

In summary, MSE is optimized for average trajectory accuracy; $L_{\mathrm{RE},\lambda}$ targets tail fidelity by exponentially upweighting extreme responses; and AWMSE offers a lightweight heuristic that emphasizes large-amplitude events through explicit weighting. We compare the learned distributions from these objectives empirically in \cref{fig:comp_pdf_134}.

\section{Identifying parametric roll in the URANS responses}

\begin{figure}[h!]
\centering

\begin{subfigure}[t]{0.32\textwidth}
  \vspace{0pt}\centering
  \includegraphics[width=0.96\linewidth]{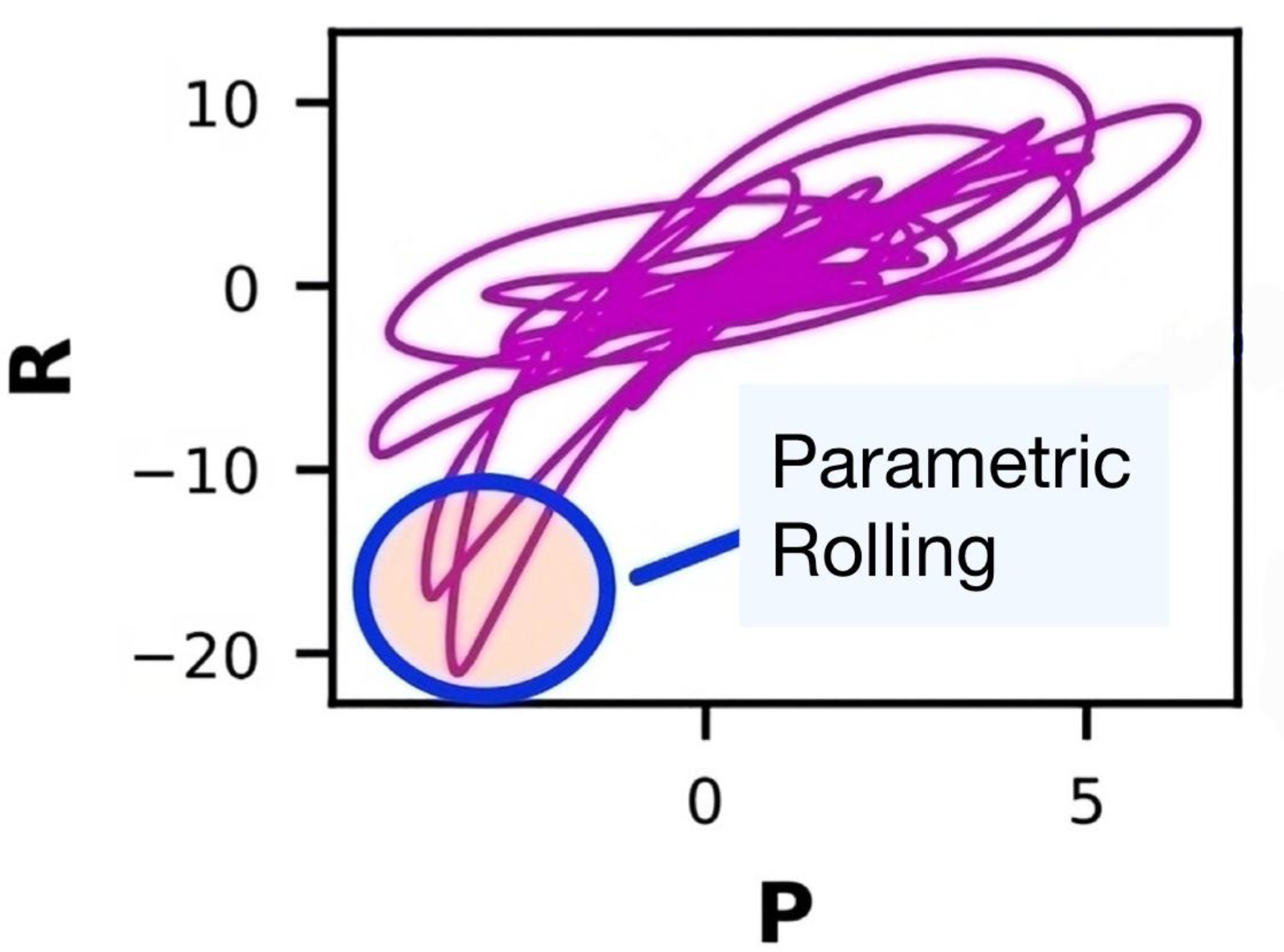}
  \caption{Characteristic roll--pitch coupling during parametric roll (phase portrait).}
  \label{fig:roll_pitch_coupling}
\end{subfigure}\hfill
\begin{subfigure}[t]{0.62\textwidth}
  \vspace{0pt}\centering
  \includegraphics[width=0.94\linewidth]{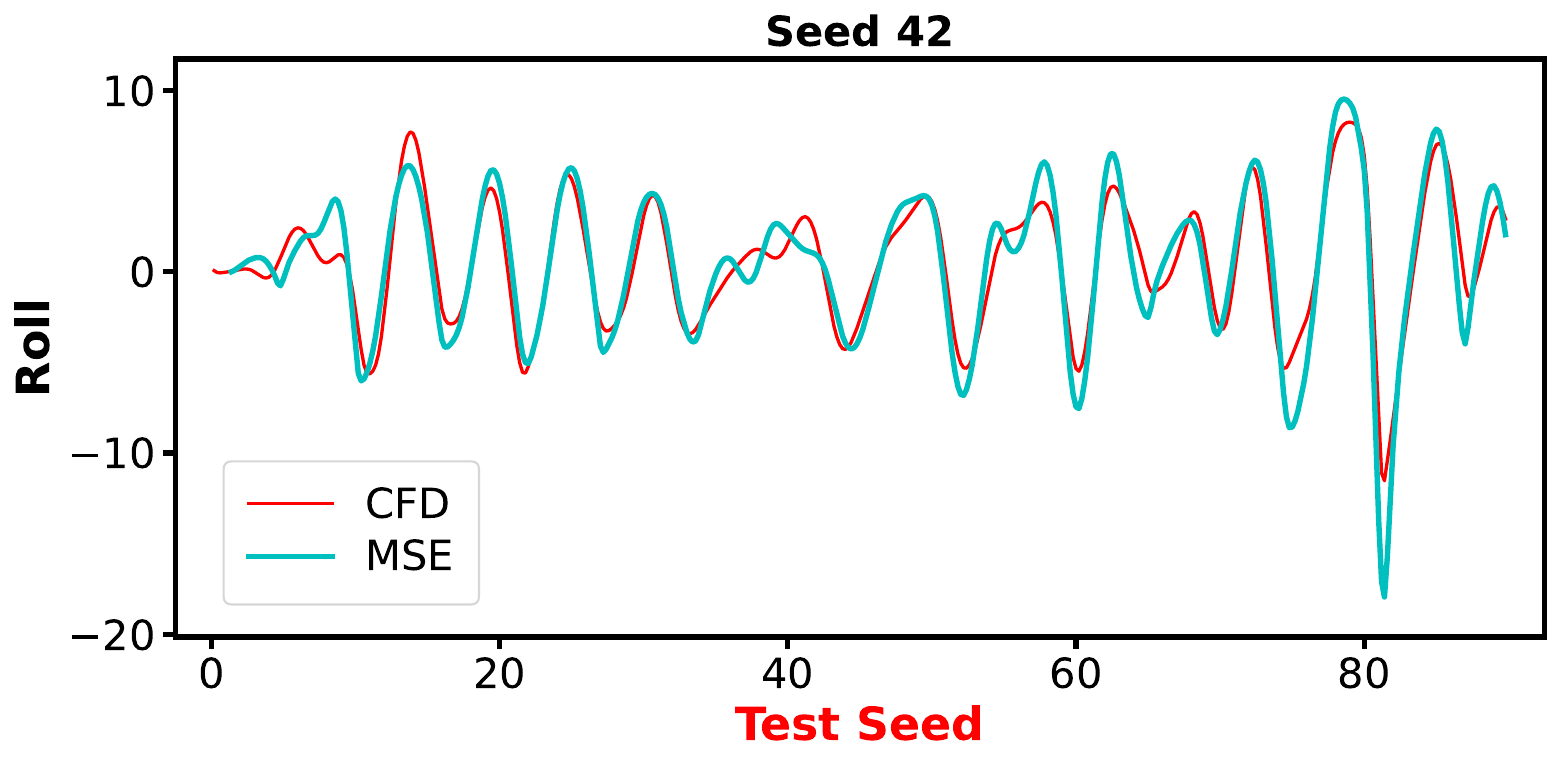}
  \caption{Representative roll prediction from prior work; \scriptsize RSE$=0.1215$.}
  \label{fig:roll_cfd_vs_lstm_prior}
\end{subfigure}

\caption{Parametric rolling already appears---and is predicted---in our earlier functional-learning study \citep{RoyProccA_JAF}, as evidenced by the roll--pitch phase portrait and CFD-versus-LSTM comparisons on held-out realizations. However, because that dataset was limited to a single sea state, it did not test whether the learned functional captures regime-dependent shifts in response statistics. Here we extend that work by pooling realizations across distinct forcing conditions and not conditioning on sea-state labels (e.g., $H_s$, $T_p$), so any changes in predicted PDFs/tails must be inferred directly from the raw wave-elevation history.}
\label{fig:parametricroll_continuity}
\end{figure}

Parametric rolling is a resonant instability that can arise under specific encounter conditions \cite{neves2006unstable,zhou2023parametricroll}. A common trigger occurs when the ship’s natural roll period is approximately twice the wave encounter period; for example, a vessel with a roll period near 12 s becomes susceptible when it repeatedly encounters waves at about 6 s. The risk is further amplified when waves approach near the bow or stern, because periodic immersion/emergence modulates buoyancy (and thus restoring) at a frequency that can lock in with the roll response \cite{france2003investigation}.

\begin{figure}[h!]
\centering
\includefig[width=0.7\linewidth]{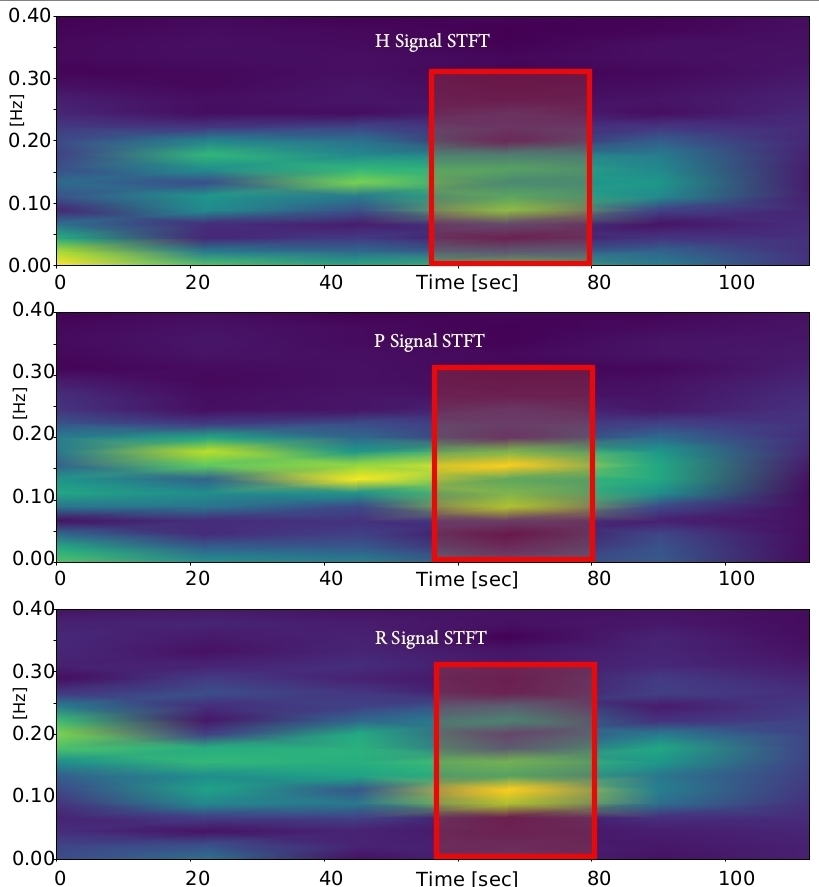}
\caption{Short-time Fourier transform (STFT) magnitude of the \textbf{H}, \textbf{P}, and \textbf{R} signals (frequency in Hz), highlighting a transition interval (red boxes; $\approx 60$--$80$ s) where energy becomes more concentrated in the response band. In the parametric-roll scenario described in the text, a vessel with $T_\phi \approx 12\,\mathrm{s}$ corresponds to $f_\phi \approx 1/T_\phi \approx 0.083\,\mathrm{Hz}$, and repeated encounters at $T_e \approx 6\,\mathrm{s}$ correspond to $f_e \approx 1/T_e \approx 0.167\,\mathrm{Hz} \approx 2 f_\phi$; the highlighted interval is consistent with the emergence/strengthening of a narrow-band component near the roll-response frequency and its associated parametric excitation. Color denotes STFT amplitude.}
\label{fig:stft}
\end{figure}

In our simulations, visualization-based estimates of encounter timing were consistent with the time–frequency diagnostics: the STFTs in \cref{fig:stft} show that the largest roll events (with a period close to 12 s) align with an encounter component near 6 s, which is a clear signature of parametric excitation. Additional evidence comes from the phase-state analysis: in the harshest sea states, the roll–pitch phase portrait in \cref{fig:roll_pitch_coupling} reveals pronounced coupling between roll and pitch, consistent with interaction with waves whose characteristic length is comparable to the vessel length and with conditions known to promote parametric rolling. Together, the STFT patterns and the roll–pitch coupling confirm that the severe-motion episodes are not simply large linear responses, but are strongly influenced by parametric rolling—highlighting the occurrence and importance of identifying these regimes for operational safety and performance.

\section{Results: episode fidelity and statistics shift}

\begin{figure*}[h!]
\centering

\begin{subfigure}[t]{0.98\textwidth}
  \centering
  \fbox{\includefig[width=\linewidth,trim=8 0 0 0,clip]{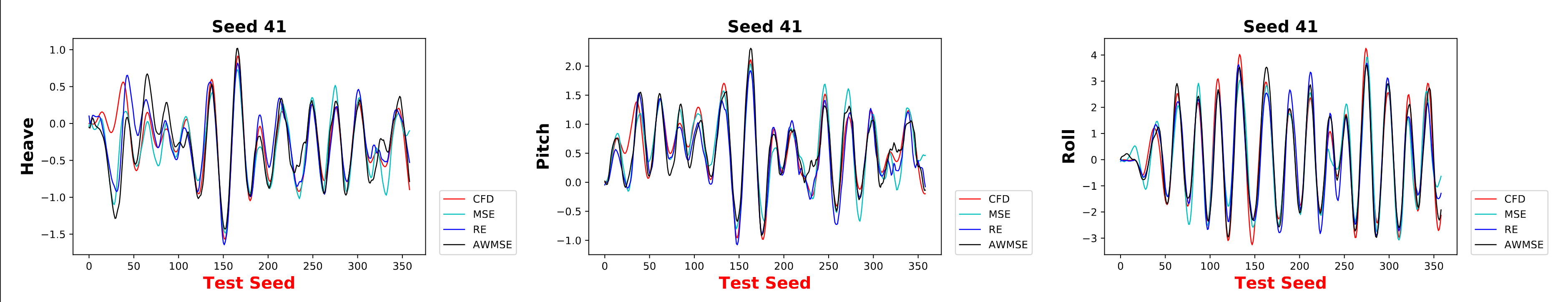}}
  \caption{SS--1 (MP97): held-out seed 41.}
  \label{fig:ts_97}
\end{subfigure}

\vspace{0.8em}

\begin{subfigure}[t]{0.98\textwidth}
  \centering
  \fbox{\includefig[width=\linewidth,trim=8 0 0 0,clip]{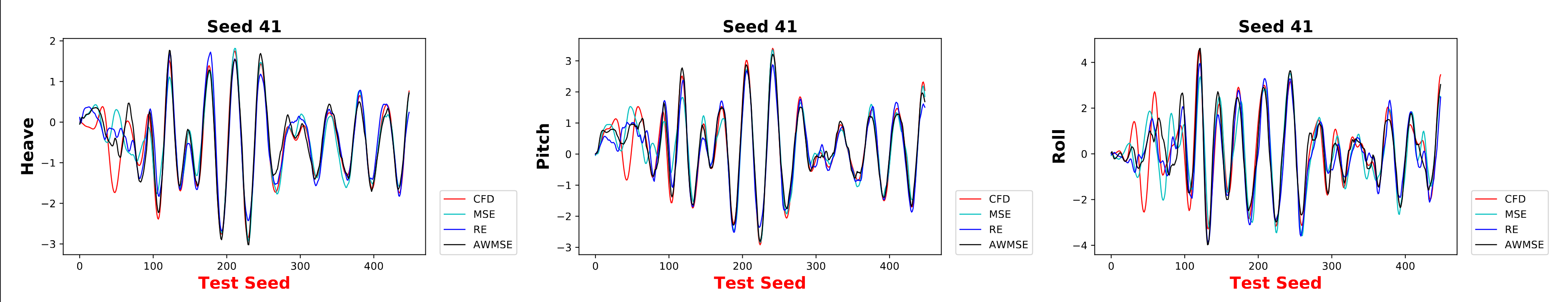}}
  \caption{SS--2 (MP124): held-out seed 41.}
  \label{fig:ts_124}
\end{subfigure}

\vspace{0.8em}

\begin{subfigure}[t]{0.98\textwidth}
  \centering
  \fbox{\includefig[width=\linewidth,trim=8 0 0 0,clip]{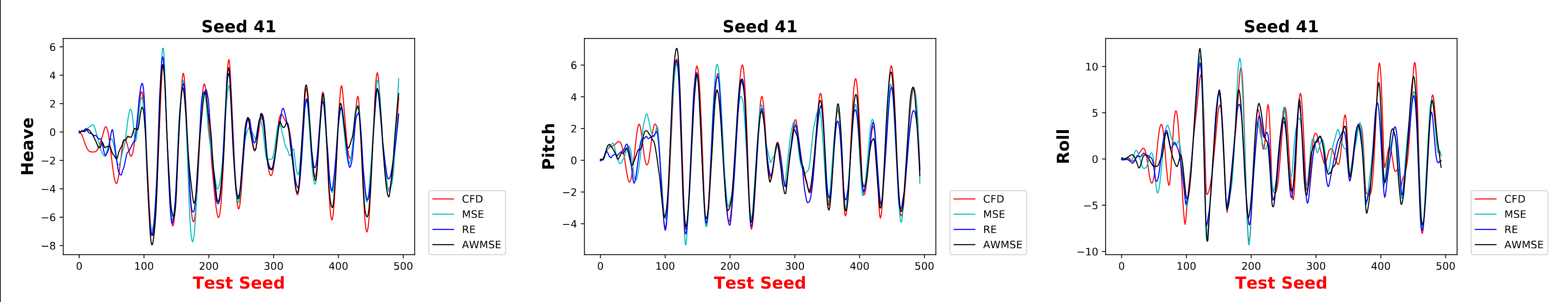}}
  \caption{SS--3 (MP134): held-out seed 41.}
  \label{fig:ts_134}
\end{subfigure}

\caption{Representative held-out time-series comparisons for seed 41 across the three sea states.
Each panel shows heave, pitch, and roll responses, comparing the URANS reference (CFD) against the learned surrogates trained with different objectives.
Across all sea states, the surrogate predictions capture the dominant oscillatory structure and timing.}
\label{fig:ts_all_seed41}
\end{figure*}

\subsection{Episode fidelity: time-series reproduction in severe seas}
We evaluate the surrogate on held-out random-phase realizations across all three sea states.
Representative time-series comparisons are shown in \cref{fig:ts_all_seed41}, where heave, pitch, and roll from the URANS reference are compared against the learned surrogates for the same held-out realization across SS--1, SS--2, and SS--3.
Across the full range of forcing conditions, the surrogate reproduces the dominant oscillatory structure and phase of the response with good fidelity.

This result should be viewed in continuity with our earlier functional-learning study \cite{RoyProccA_JAF}, summarized in \cref{fig:parametricroll_continuity} and also used in this paper.
Although not identified, that earlier work already showed that LSTM-based functionals can reproduce parametric-roll-like episodes in time domain for held-out realizations within a single sea state.
The present results extend that finding in two important ways: first, the same functional formulation is now evaluated across multiple sea states with distinct response regimes; second, the model is not conditioned on explicit sea-state labels, so the episode-level behavior must be inferred directly from the raw wave-elevation history.
Accordingly, \cref{fig:ts_all_seed41,fig:parametricroll_continuity} together show that the surrogate is not merely fitting local oscillations, but is able to follow the emergence and development of nonlinear roll episodes under increasingly severe forcing.

\subsection{Statistics shift: roll PDF fidelity and tail behavior}
For parametric-roll assessment, reproducing the time series alone is not sufficient: the critical question is whether the surrogate also captures the \emph{change in response statistics} associated with the onset of the nonlinear regime.
This is precisely the gap left by the earlier single-sea-state study in \cref{fig:parametricroll_continuity}: while time-domain parametric-roll episodes could already be tracked, that dataset could not test whether the learned functional preserved regime-dependent shifts in the roll distribution.

The present results address that question directly.
\Cref{fig:comp_pdf_134} compares the URANS roll PDF in the most severe sea state (SS--3) against the PDFs induced by surrogates trained under different loss functions.
The comparison shows that loss design materially affects the fidelity of the predicted tail, and therefore the surrogate’s usefulness for risk-sensitive applications.
In particular, the heavy-tail structure associated with intermittent large-roll episodes can be better preserved when the objective places additional emphasis on extreme responses, whereas standard MSE tends to prioritize average trajectory fidelity over low-probability tail behavior.
Taken together with the time-domain evidence in \cref{fig:ts_all_seed41}, these results show that the relevant notion of success in severe seas is \emph{joint}: the surrogate must reproduce both the event-level dynamics and the accompanying distributional shift in the roll response.


\begin{figure*}[t]
\centering

\begin{subfigure}[t]{0.32\textwidth}
  \vspace{0pt}\centering
  \includefig[width=\linewidth]{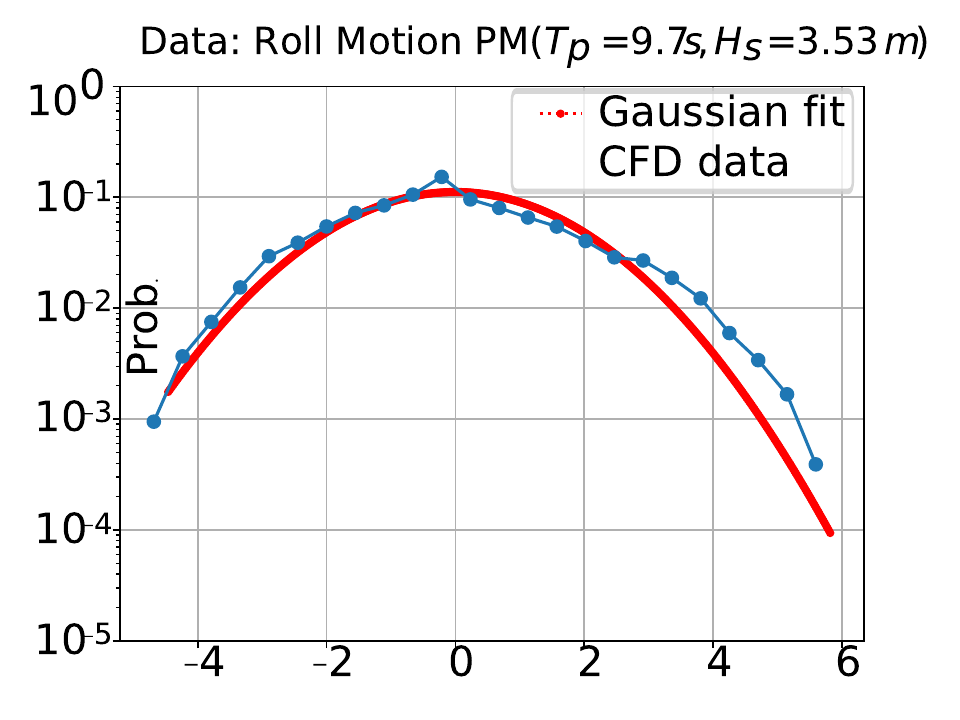}
  \caption{SS--1}
  \label{fig:roll_pdf_ss1}
\end{subfigure}\hfill
\begin{subfigure}[t]{0.32\textwidth}
  \vspace{0pt}\centering
  \includefig[width=\linewidth]{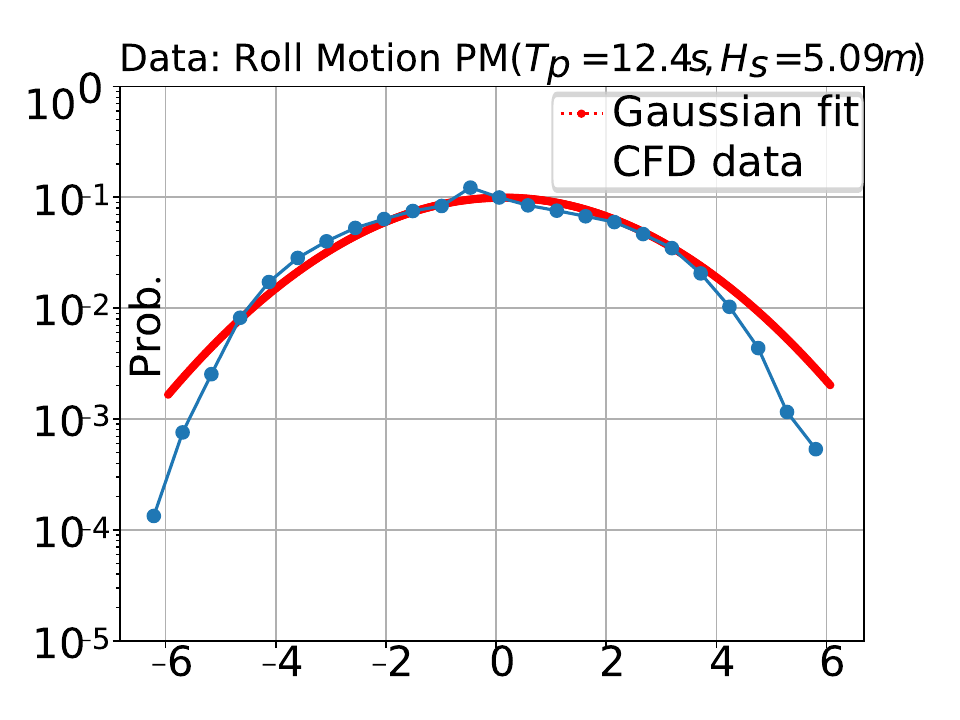}
  \caption{SS--2}
  \label{fig:roll_pdf_ss2}
\end{subfigure}\hfill
\begin{subfigure}[t]{0.32\textwidth}
  \vspace{0pt}\centering
  \includefig[width=\linewidth]{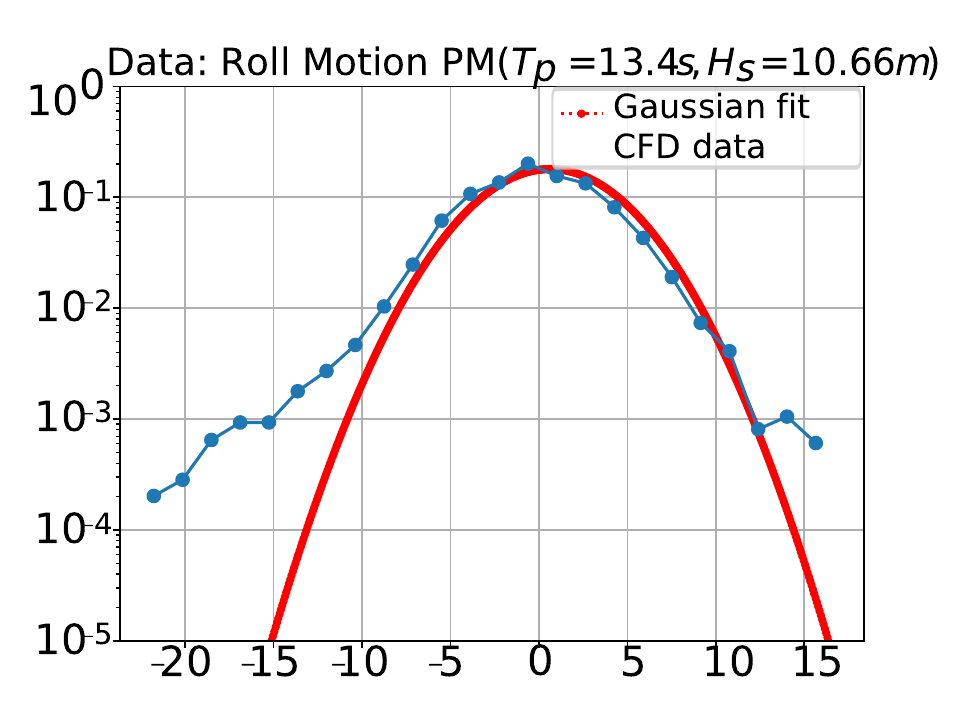}
  \caption{SS--3}
  \label{fig:roll_pdf_ss3}
\end{subfigure}

\caption{Reference roll response statistics across sea states.
Each panel shows the predicted roll probability density (PDF) for a held-out ensemble in SS--1/SS--2/SS--3.
The distribution remains approximately Gaussian in SS--1 and SS--2, while SS--3 exhibits pronounced non-Gaussian tail behavior consistent with intermittent large-roll episodes (parametric-roll regime).}
\label{fig:roll_pdf_shift_across_seastates}
\end{figure*}


\section{Discussion}

The defining operational risk in parametric roll is not that a predictor is slightly wrong in a mean-square sense, but that it misrepresents the probability of \emph{rare large responses}.
This is visible directly in the cross-sea-state behavior: \Cref{fig:roll_pdf_shift_across_seastates} shows that the reference roll distribution remains close to Gaussian in SS--1/SS--2, yet develops a pronounced non-Gaussian tail in SS--3.
That separation of regimes is the core practical requirement for operability assessment, because the decisions of interest (limits, thresholds, exceedance probabilities) depend on tail mass rather than on typical oscillations near the mode.


The URANS ensemble provides independent evidence that the SS--3 heavy-tail behavior corresponds to parametric excitation rather than merely ``bigger linear response.''
The time--frequency diagnostic in \Cref{fig:stft} is consistent with the canonical subharmonic condition (encounter component near twice the dominant roll-response frequency), and the roll--pitch phase portrait in \Cref{fig:roll_pitch_coupling} shows the characteristic coupling associated with parametric-roll dynamics.
These two signals together justify interpreting the SS--3 PDF deformation in \Cref{fig:roll_pdf_shift_across_seastates} as a \emph{regime change} in the underlying dynamics and response statistics.

Parametric roll episodes were already present—and could be tracked in time domain—in our earlier functional-learning study \citep{RoyProccA_JAF}, as summarized in \Cref{fig:parametricroll_continuity}.
What that earlier dataset could not test was whether a learned wave-history$\rightarrow$motion functional also preserves \emph{regime-dependent response statistics}, because training and evaluation were confined to a single sea state.
The present study closes that gap by pooling realizations across distinct forcing conditions and \emph{not} providing sea-state labels (e.g., $H_s$, $T_p$) as inputs: the regime separation and tail changes in \Cref{fig:roll_pdf_shift_across_seastates} must therefore be inferred from raw elevation histories alone.
This ``no-regime-label'' setting makes the statistics-shift result a stronger test of functional learning in severe seas, and directly matches the use case where only local wave measurements are available.

The comparison in \Cref{fig:comp_pdf_134} shows that loss-function choice can materially change the surrogate-induced roll PDF in the most severe regime.
MSE provides a strong baseline for average trajectory fidelity, but the tail-sensitive deviations that drive risk can remain underemphasized when extreme events occupy a small fraction of samples.
In contrast, the relative-entropy-inspired objective and amplitude-weighted variants explicitly reweight the learning signal toward large-amplitude outcomes, and this reweighting can translate into improved agreement in tail regions and better qualitative preservation of the non-Gaussian structure in SS--3.
Practically, this means the surrogate can be tuned depending on whether the downstream objective is average time-domain accuracy or tail-risk assessment.



\begin{figure*}[t!]
\centering

\begin{subfigure}[t]{0.48\textwidth}
    \centering
    \includegraphics[width=\textwidth]{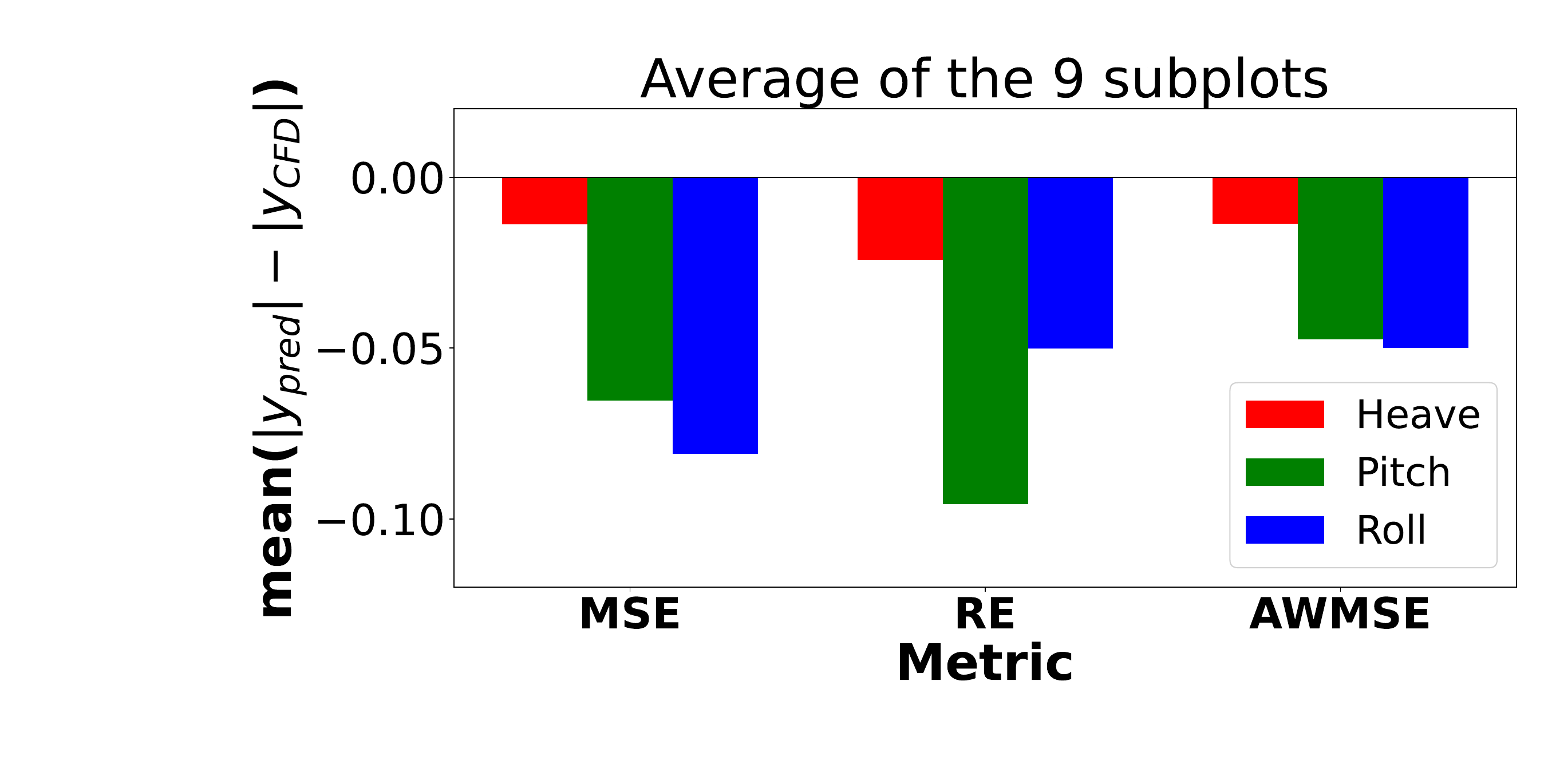}
    \caption{Sea State 1}
    \label{fig:lstm_avg_ss1}
\end{subfigure}\hfill
\begin{subfigure}[t]{0.48\textwidth}
    \centering
    \includegraphics[width=\textwidth]{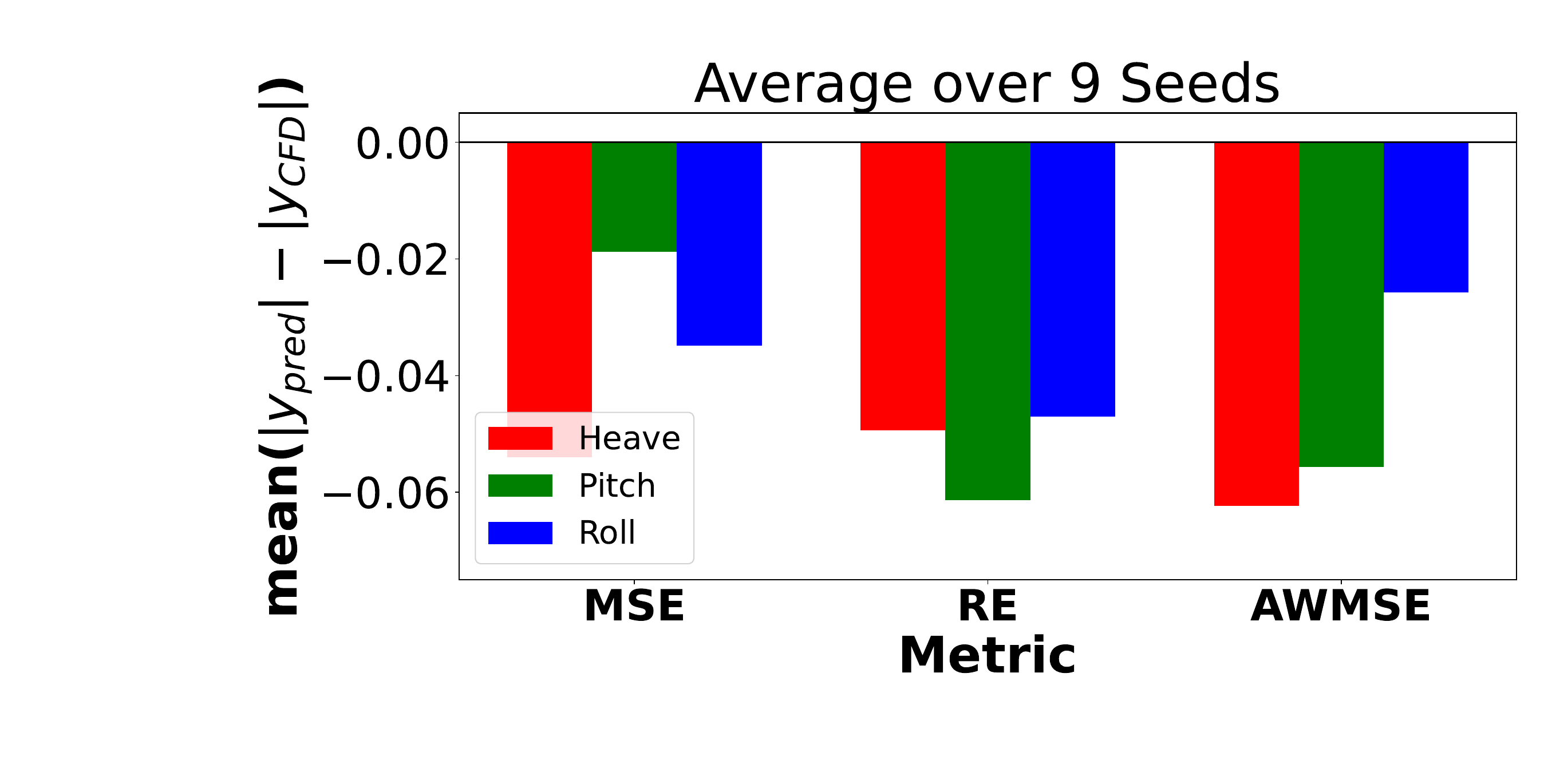}
    \caption{Sea State 2}
    \label{fig:lstm_avg_ss2}
\end{subfigure}

\vspace{0.8em}

\begin{subfigure}[t]{0.48\textwidth}
    \centering
    \includegraphics[width=\textwidth]{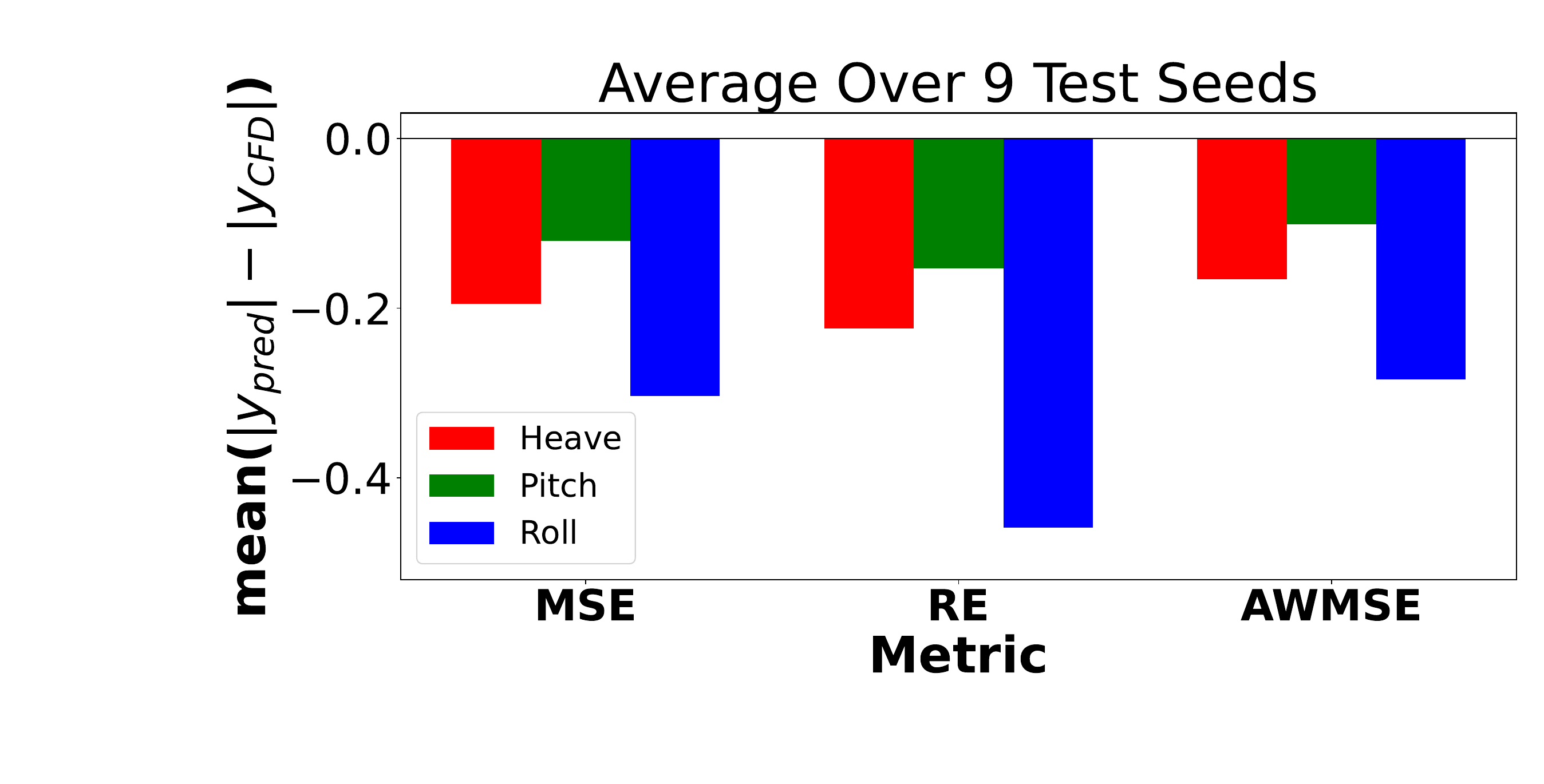}
    \caption{Sea State 3}
    \label{fig:lstm_avg_ss3}
\end{subfigure}

\caption{\textit{Averaged LSTM test results} (6 hidden layers, 250 neurons). 
Bar charts show the average of the Heave, Pitch, and Roll absolute-difference metrics 
(MSE, RE, and AWMSE) over the 9 testing seeds for a notional DTMB battleship at 
Froude number 0.4, time step $\Delta t = 0.2\,\mathrm{s}$, and speed 
$v = 30\,\mathrm{kt}$, for Sea States 1, 2, and 3.}
\label{fig:lstm_avg_bar_seastates}
\end{figure*}

\section{Conclusions}

We developed an end-to-end, data-source-agnostic recurrent functional surrogate that maps wave-elevation histories to vessel motions and evaluated it in a severe-sea URANS ensemble spanning three sea states.
The results show that the surrogate reproduces the dominant oscillatory structure and timing of the responses across all sea states, including time-domain large-roll episodes consistent with parametric excitation \cref{fig:ts_all_seed41}.
This is consistent with our earlier single-sea-state functional-learning study, which already showed that parametric rolling and the associated roll--pitch coupling could be reproduced on held-out realizations \cref{fig:roll_pitch_coupling,fig:roll_cfd_vs_lstm_prior}.
The key advance here is that, by moving beyond a single forcing condition and not conditioning on explicit sea-state labels, we show that the same functional framework can also recover regime-dependent changes in response statistics directly from raw wave-elevation histories.
Its main deficiency is not a breakdown of the underlying temporal dynamics, but a systematic underprediction \cref{fig:lstm_avg_bar_seastates} of motion amplitude, which becomes most visible in the most severe sea state and is strongest in roll.

Even so, the most severe sea state should not be read simply as a failure case.
Part of the increase in error reflects the fact that the underlying motions are substantially larger in harsher conditions.
More importantly, the surrogate still captures the accompanying regime-dependent shift in roll response statistics, including the pronounced departure from Gaussian behavior in the most severe sea state.
The PDF comparisons, \cref{fig:comp_pdf_134} are especially encouraging in this regard: although trajectory-level errors increase, the learned roll distributions remain much closer to the URANS statistics than a Gaussian fit, showing that the model preserves the essential non-Gaussian character of the severe-state response.

Crucially, these statistical shifts are learned without conditioning on explicit sea-state labels, implying that the functional model can infer regime changes directly from raw wave-history inputs.
Finally, the PDF comparisons demonstrate that loss-function design provides a controllable tradeoff between average trajectory fidelity and tail fidelity.
Among the objectives considered, the amplitude-weighted MSE provides the most balanced and robust overall performance across sea states and motion components, supporting the use of risk-aware recurrent surrogates for rapid operability and exceedance-style assessments at a fraction of CFD cost.
\section*{Funding}

The author was supported by the ``la Caixa'' Fellowship [award number LCF/BQ/AA17/1161000].
The funding source had no involvement in the study design; in the collection,
analysis, or interpretation of data; in the writing of the report; or in the
decision to submit the article for publication.

\section*{Data Availability}

The URANS wave--motion dataset supporting this study has been deposited at
\url{https://github.com/j-aguila/MotionsDTMB3Spectra}. The dataset will be made
publicly available upon publication of this article. During peer review, access
may be provided to the editor and reviewers upon reasonable request.

\section*{Declaration of generative AI and AI-assisted technologies in the manuscript preparation process}

During the preparation of this work, the author used ChatGPT (OpenAI) to assist
with language editing, improving readability, and refining the presentation of
the manuscript text. After using this tool, the author reviewed and edited the
content as needed and takes full responsibility for the content of the published
article.

\section*{Acknowledgements}

The author would like to sincerely thank Professors Michael Triantafyllou,
Themis Sapsis, and George Karniadakis for their intellectual influence on the
broader research path from which this work emerged. Professor Triantafyllou's
guidance was especially important in shaping the ocean-engineering perspective
and motivation behind the study. The author is also grateful for Professor
Sapsis's influence on thinking around nonlinear response, rare events, and
statistical transitions, and for Professor Karniadakis's influence on
scientific machine learning and the use of data-driven methods for complex
physical systems.

\bibliographystyle{unsrtnat}
\bibliography{main}

\end{document}